\documentclass{article}
\pdfoutput=1
\usepackage{PRIMEarxiv}
\usepackage{amsmath,amssymb,amsfonts}
\usepackage{algorithmic}
\usepackage{algorithm}
\usepackage{graphicx}
\usepackage{booktabs}
\usepackage{tablefootnote}
\usepackage{threeparttable}
\usepackage{tabularx}  
\usepackage{adjustbox}  
\usepackage[switch, pagewise]{lineno}  
\usepackage[authoryear]{natbib}
\usepackage{stfloats}  
\usepackage{hyperref}

\begin{document}
\title{Weight-Freezing: A Regularization Approach for Fully Connected Layers with an Application in EEG Classification}

\author{
    Zhengqing Miao \footnotemark[1] \And
    Meirong Zhao   \footnotemark[1] \And
    }
\renewcommand{\thefootnote}{\fnsymbol{footnote}}
\footnotetext{Corresponding author: Zhengqing Miao (mzq@tju.edu.cn)}
\footnotetext[1]{State Key Laboratory of Precision Measuring Technology and Instruments, School of Precision Instrument and Opto-electronics Engineering, Tianjin University, Tianjin 300072, China.}

\maketitle

\begin{abstract}

In the realm of EEG decoding, enhancing the performance of artificial neural networks (ANNs) carries significant potential. This study introduces a novel approach, termed "Weight-Freezing", that is anchored on the principles of ANN regularization and neuroscience prior knowledge. The concept of Weight-Freezing revolves around the idea of reducing certain neurons' influence on the decision-making process for a specific EEG task by freezing specific weights in the fully connected layer during the backpropagation process.
This is actualized through the use of a mask matrix and a threshold to determine the proportion of weights to be frozen during backpropagation. Moreover, by setting the masked weights to zero, Weight-Freezing can not only realize sparse connections in networks with a fully connected layer as the classifier but also function as an efficacious regularization method for fully connected layers.
Through experiments involving three distinct ANN architectures and three widely recognized EEG datasets, we validate the potency of Weight-Freezing. Our method significantly surpasses previous peak performances in classification accuracy across all examined datasets. Supplementary control experiments offer insights into performance differences pre and post Weight-Freezing implementation and scrutinize the influence of the threshold in the Weight-Freezing process.
Our study underscores the superior efficacy of Weight-Freezing compared to traditional fully connected networks for EEG feature classification tasks. With its proven effectiveness, this innovative approach holds substantial promise for contributing to future strides in EEG decoding research.


\end{abstract}

\section{Introduction}
\label{sec:introduction}

Electroencephalography (EEG), due to its non-invasive nature, high temporal resolution, portability, and cost efficiency, has been widely applied in Brain-Computer Interface (BCI) systems (\cite{wolpaw2007brain, michel2012towards}). For instance, Motor Imagery (MI) signals (\cite{lotze2006motor}), a subset of EEG signals generated when an individual imagines performing a motor task without actual physical movement, are particularly interesting for BCI applications as they can be used in motor rehabilitation and brain function regulation (\cite{pfurtscheller2001motor, park2014assessment, sterman1996physiological}).
However, accurate identification of brain activity corresponding to EEG signals is challenging due to its low spatial resolution, low signal-to-noise ratio, and the non-stationary and inherent variability in brain activity (\cite{kaplan2005nonstationary, goncharova2003emg}).
Common Spatial Patterns (CSP) and related methods (\cite{lotte2010regularizing, ang2011filter}) are machine learning techniques commonly used for feature extraction in MI. They identify optimal channels and features from EEG signals most relevant to a particular task. Among these, Filter Bank Common Spatial Patterns (FBCSP) has played a significant role in MI signal feature extraction by applying a set of bandpass filters to the raw EEG data, generating spectrum-specific signals for each filter band (\cite{ang2008filter}).

Given the powerful feature extraction and classification capabilities of Artificial Neural Networks (ANNs), they are becoming a popular choice for decoding EEG signals in BCI applications (\cite{schwemmer2018meeting, acharya2018deep}). For example, Schirrmeister et al. (2017) explored the feature extraction capabilities of Shallow-ConvNet and Deep-ConvNet for MI and Motor Execution (ME) EEG signals (\cite{schirrmeister2017deep}). Lawhern et al. (2018) expanded on the Shallow-ConvNet decoder by adding a temporal convolution layer and using separable convolutions to enhance the decoder's performance across various EEG paradigms (\cite{lawhern2018eegnet}). Borra et al. (2020) proposed a lightweight shallow CNN, which stacks a temporal sinc-convolutional layer and a spatial depthwise convolutional layer to extract efficient MI- and ME-EEG features (\cite{borra2020interpretable}). In our previous work, we proposed the LMDA-Net, which added a channel attention module and depth attention module to the original temporal and spatial convolutions to enhance the feature extraction capabilities for various BCI tasks (\cite{miao2023lmda}). These models are all end-to-end artificial neural networks that aim to enhance ANN decoding of EEG signals from the perspective of feature extraction networks. However, to the best of our knowledge, no work has studied the impact of the classifier in end-to-end ANNs on EEG decoding performance. One important reason might be that the setup of classifiers in machine vision, natural language processing, or EEG decoding is relatively fixed, usually using one or several fully connected layers for classification. The question this research faces is whether the existing fully connected network is the optimal classifier for EEG, a signal with low signal-to-noise ratio and small data volume.

\begin{figure*}
\centerline{\includegraphics[width=0.9\textwidth]{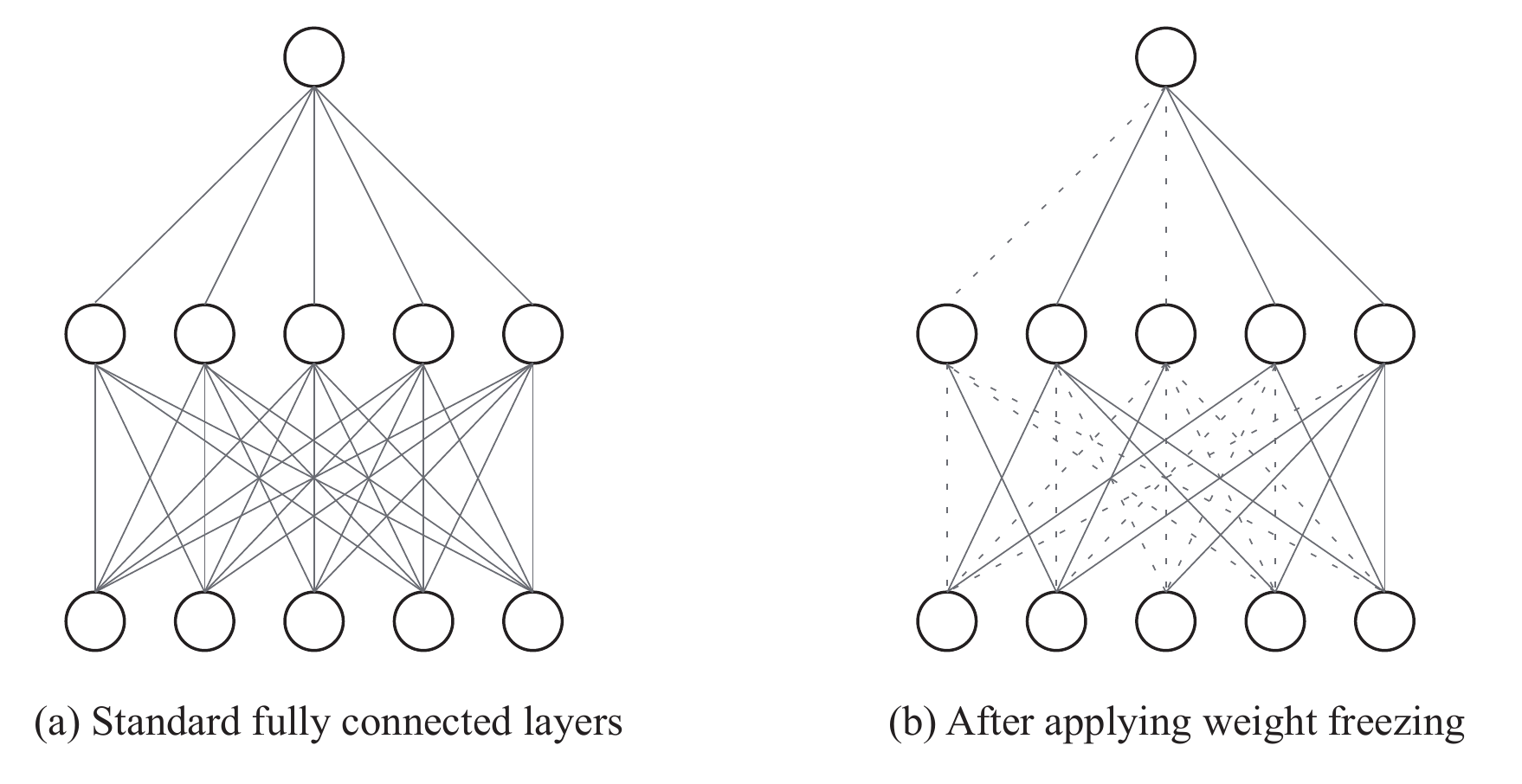}}
\caption{The comparison diagram before and after Weight-Freezing. (a) Standard fully connected network. (b) Fully connected layer with Weight-Freezing. The dashed lines represent the frozen weights.
}
\label{fig1}
\end{figure*}

To investigate this question, this study proposes a Weight-Freezing technique. As the name suggests, Weight-Freezing freezes some weights in the backpropagation process of the fully connected layer. As shown in Figure \ref{fig1}, compared with the fully connected network, Weight-Freezing suppresses the update of some parameters in the fully connected network, thereby effectively suppressing the influence of some input neurons on the decision result during the classification decision process.

The major contributions  of this paper are as follows:
\begin{enumerate}
    \item To the best of our knowledge, this paper is the first to study the impact of the classifier in ANNs on EEG decoding performance. For this purpose, Weight-Freezing is proposed, which suppresses the influence of some input neurons on certain decision results by freezing some parameters in the fully connected layer, thereby achieving higher classification accuracy.
    \item Weight-Freezing is also a novel regularization method, which can achieve sparse connections in the fully connected network.
    \item Weight-Freezing is thoroughly validated and analyzed in three classic decoding networks and three highly cited public EEG datasets. The experimental results confirm the superiority of Weight-Freezing in classification and have also achieved state-of-the-art classification performance (averaged across all participants) for all the three highly cited datasets.
\end{enumerate}

This study's primary contribution lies in its potent facilitation of the application and implementation of Artificial Neural Network (ANN) models within Brain-Computer Interface (BCI) systems. Simultaneously, it sets a new performance benchmark for future EEG signal decoding efforts using more sizable models, such as transformers \cite{vaswani2017attention}.
Emerging research (\cite{ahn2022multiscale, bagchi2022eeg, zhang2023mi, ma2023mbga, song2023Conformer}) is increasingly adopting transformer networks for EEG signal decoding. These approaches can be viewed as enrichments to existing ANN models, as they elevate EEG classification accuracy via more sophisticated feature extraction networks. However, these enhancements have inadvertently complicated the deployment of these ANN models in real-world BCI systems.
In a stark contrast, our study introduces Weight-Freezing as an innovative, subtractive strategy that refines existing ANN models. Empowered by Weight-Freezing, some lightweight and shallow decoding networks surpass all current transformer-based methods in terms of classification performance on identical public datasets.
The incorporation of Weight-Freezing not only simplifies the deployment of ANN models within BCI systems but also sets a new performance standard for the deployment of larger models, such as transformers, in the future. Moreover, it provokes an intriguing question in the realm of EEG decoding: Is the deployment of large models like transformers for EEG feature extraction truly indispensable?

This rest of this paper is structured as follows. Section \ref{sec:motivation} provides a detailed introduction to the motivation behind the introduction of Weight-Freezing from three perspectives. Section \ref{sec:method} explains the principle of Weight-Freezing from the perspective of information propagation in ANN models. Section \ref{sec:datasets} introduces the dataset used in this study. In Section \ref{sec:Experiments}, a detailed analysis is conducted on the classification performance of three classic decoding networks with the support of Weight-Freezing on three highly cited EEG public datasets. Section \ref{sec:analyses} performs rigorous comparative experiments to analyze the differences and robustness between Weight-Freezing and fully connected networks. Lastly, Section \ref{sec:Conclusion-outlook} summarizes the work presented in this paper and provides an outlook on future research directions.

\section{Motivation}
\label{sec:motivation}

\subsection{Decoding Capability of ANNs}

When using ANNs for EEG signal classification, previous works have primarily focused on designing feature extraction networks. For example, Shallow-ConvNet (\cite{schirrmeister2017deep}) employed temporal and spatial convolutional networks based on the characteristics of EEG signals for feature extraction. ANN models like EEGNet (\cite{lawhern2018eegnet}), DRDA \cite{zhao2020deep} and Conformer (\cite{song2023Conformer}) believed that the feature extraction capability of these two layers is limited, leading them to deepen the layers of the feature extraction network. Sinc-ShallowNet (\cite{borra2020interpretable}) and LMDA-Net (\cite{miao2023lmda}), on the other hand, horizontally expanded the feature extraction capability of temporal and spatial convolution without deepening the layers of the feature extraction network. Undeniably, designing EEG-specific feature extraction networks is crucial for decoding EEG signals. However, in end-to-end ANN models, where the feature extraction network and the classifier form an integrated system, the optimality of fully connected networks for EEG feature classification remains a question.

This study takes a different perspective by focusing on enhancing the classification capability of the classifier to address the EEG classification problem. The fundamental assumption of this study is that the features input into the classifier are mixed with noise, which arises from the low signal-to-noise ratio and limited data availability in EEG. A low signal-to-noise ratio implies the presence of widespread noise in the signal, while limited data availability hampers the efficient training of the feature extraction layer in the ANN, thereby hindering the elimination of noise's impact on feature extraction and classification.

\subsection{Overfitting in ANNs}

In end-to-end ANNs, the feature extraction network and the classification network form an integral entity. From the perspective of forward and backward propagation, the feature extraction network and the classification network complement each other. Models like EEGNet (\cite{lawhern2018eegnet}), Sinc-ShallowNet (\cite{schirrmeister2017deep}), and LMDA-Net (\cite{miao2023lmda}) all employ separable convolutions in the feature extraction network, enabling sparse connections within the feature extraction network. This design effectively addresses the characteristics of limited data volume and low signal-to-noise ratio in EEG while effectively avoiding overfitting of the neural network model.

Although the ANNs used for EEG decoding have significant differences in the design of the feature extraction layer, Shallow-ConvNet, Deep-ConvNet, EEGNet, Sinc-ShallowNet, and LMDA-Net all use a fully connected layer for classification. Because the fully connected network is densely connected, that is, when the input and output of the fully connected layer are $n_1$ and $n_2$ respectively, the fully connected layer needs to introduce $n_{1}\times n_2$ learnable parameters. For EEG data with only a few hundred training samples, this can easily cause overfitting (\cite{bejani2021systematic, santos2022avoiding, devries2017improved}). This is also the reason why the above classic ANNs only use a single fully connected network for classification. To minimize overfitting in the fully connected layer, LMDA-Net uses three strategies to reduce the number of neurons input to the fully connected layer as much as possible, namely using a small number of spatial convolution kernels in the spatial convolution layer (the layer before the fully connected layer), adaptive pooling, and a large dropout rate (P=0.65).

However, the above down-sampling or regularization methods all act before the fully connected layer, and there is currently no effective regularization method to constrain the dense connection  of the fully connected layer itself. If the dense connection method of the fully connected layer is changed to sparse connection, it will greatly reduce the number of learnable parameters in the fully connected layer, which may be an advantage for EEG signals with a small amount of data and a low signal-to-noise ratio.

\subsection{Prior Knowledge in Neuroscience}

Using fully connected layers for EEG classification implies that each neuron in the fully connected layer contributes to the decision-making process for a specific neural activity category. This means that the judgment of a specific neural activity requires the involvement of all input neurons.
However, this decision-making process does not align with prior knowledge from neuroscience. For example, in motor imagery tasks, activating the left-hand motor imagery task triggers the motor cortex in the right hemisphere, while activating the right-hand motor imagery task triggers the motor cortex in the left hemisphere. Similarly, activating the two-foot motor imagery task triggers the central area of the brain's motor cortex (\cite{pfurtscheller2006mu, pfurtscheller2001motor}). This prior knowledge suggests that when determining the category of motor imagery, it is not necessary to consider the electrical signals from all neurons. Instead, a classifier that aligns more with neuroscience's prior knowledge should only take into account the impact of specific input neurons on the category during the determination process. In other words, employing a sparse connection method within the fully connected layer better aligns with neuroscience's prior knowledge.

\section{Method}
\label{sec:method}

\subsection{Principle of Weight-Freezing}

Taking a single fully connected layer as an example, let's now explore the specific principles and implementation methods of Weight-Freezing.
Suppose the features input into the fully connected layer are $\mathbf{X}=\{\mathbf{x}_{1},\mathbf{x}_{2}, \cdot, \mathbf{x}_n\}$, where $\mathbf{X}\in \mathbb{R}^{N \times{L}}$. Here, $N$ is the size of the mini-batch, and $L$ denotes the dimension of the feature. Let $n \in\{1, N\}$ index the mini-batch of the network. $\mathbf{x}_n$ represents the vector of inputs into the fully connected layer, and $\mathbf{y}_n$ denotes the vector of output from the fully connected layer. $\mathbf{W}_n$ and $\mathbf{b}_n$ are the weights and biases of the fully connected layer. The feed-forward operation of the fully connected layer can be described as: 
\begin{equation} \label{eq1}
\mathbf{y}_{n}=\mathbf{W}_n\mathbf{x}_{n}+\mathbf{b}_n
\end{equation}

Assuming that the output of the fully connected layer is directly classified through Cross Entropy Loss, and the output of Cross Entropy Loss is denoted as $\mathcal{L}$, $z_n$ represents the target, and $C$ is the number of classes. The output of Cross Entropy Loss can be described as :
\begin{equation} \label{eq2}
\mathcal{L}=-\sum_{n=1}^{N} \log \frac{e^{\mathbf{y}_{n, z_n}}}{\sum_{c=1}^{C} e^{\mathbf{y}_{n, c}}}
\end{equation}

In equation \ref{eq1}, $\mathbf{W}_n$ and $\mathbf{b}_n$ are learnable parameters that undergo iterative adjustments guided by the gradients computed through the backpropagation algorithm. The derivative of the learnable parameter $\mathbf{W}_n$ in the training process is shown in equation \ref{eq3}:
\begin{equation} \label{eq3}
\frac{\partial \mathcal{L}}{\partial W_{n}} = \frac{\partial \mathcal{L}}{\partial y_{n}} \cdot \frac{\partial y_{n}}{\partial W_{n}} = (\tilde{z}_n -z_n)\cdot\mathbf{x}_n^T
\end{equation}
Here, $\tilde{z}_n$ represents the predicted value output by the softmax function in CrossEntropyloss.  \footnote{The derivation process of Cross Entropy loss is widely available online, so the specific derivation process is omitted here.}

Let $\eta$ represent the learning rate of the optimizer, then the update of the parameter $\mathbf{W}_n$ in the training process of the neural network model can be expressed as:
\begin{equation}
\mathbf{W}_n=\mathbf{W}_n-\eta\cdot(\tilde{z}_n -z_n)\mathbf{x}_n^T
\end{equation}

The above is a complete process of the role of the weight $\mathbf{W}_n$ in the forward propagation and the update process in the backpropagation of a fully connected layer. The Weight-Freezing method does not change the forward propagation process of the fully connected layer, but in the backpropagation process, as the name suggests, it freezes some of the learnable parameters in the fully connected layer, preventing them from being updated through back. The specific implementation of this method can be represented by equation \ref{eq5} :
\begin{equation} \label{eq5}
\mathbf{W}_n = \mathbf{W}_n-\mathbf{M}\odot (\eta\cdot(\tilde{z}_n -z_n)\mathbf{x}_n^T))
\end{equation}
In equation \ref{eq5}, $\mathbf{M}$ is a mask matrix with the same dimensions as $\mathbf{W}_n$, where the elements follow a uniform distribution in the range [0, 1). $t$ ($0 \leq t \leq 1$) represents the threshold of the mask matrix, where a larger $t$ indicates a higher degree of masking, resulting in more parameters being frozen in $\mathbf{W}_n$. The $\odot$ symbol denotes element-wise multiplication, also known as Hadamard product. If an element is masked, it means that it is frozen during backpropagation and cannot be updated through backpropagation.

Moreover, Weight-Freezing can also facilitate sparse connections in the fully connected layer. By manually setting the masked weights to zero in the mask matrix $\mathbf{M}$, the masked portion becomes inactive during both forward and backward propagation, achieving sparse connections in the fully connected layer.

\subsection{Difference from Dropout}

The Dropout (\cite{srivastava2014dropout, wager2013dropout, wu2015towards}) method can be mathematically represented as equation \ref{eq6}:
\begin{equation} \label{eq6}
\mathbf{v}_\text{{dropout}}=\mathbf{v}\odot \mathbf{m}
\end{equation}
In equation \ref{eq6}, $\mathbf{v}$ denotes the input vector of the Dropout layer, and $\mathbf{m}$ is a mask vector with the same dimensions as $\mathbf{v}$. The elements of $\mathbf{m}$ are independently drawn from a Bernoulli distribution with parameter $p$. If an element is drawn, the corresponding neuron will be "dropped," meaning its output is set to 0.

\begin{figure*}
\centerline{\includegraphics[width=0.9\textwidth]{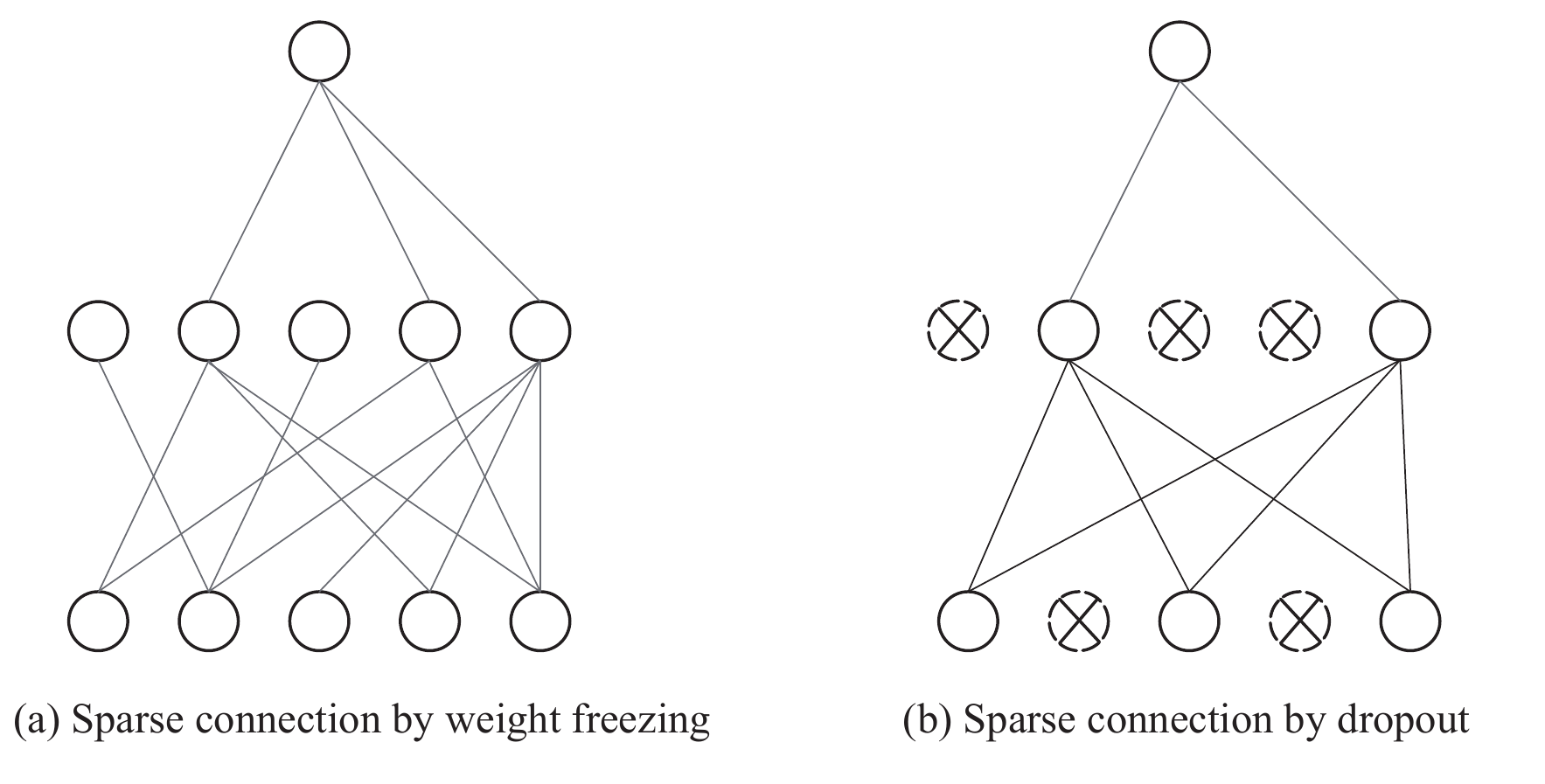}}
\caption{The distinction between dropout and Weight-Freezing in regularization}
\label{fig2}
\end{figure*}

Both Weight-Freezing and Dropout can be considered as regularization methods to prevent overfitting in neural networks. The comparison of the implementation principles of Weight-Freezing and Dropout is as follows:
\begin{enumerate}
    \item \textbf{Target of Action}: As can be seen from Figure \ref{fig2}, dropout operates by altering the neurons, while Weight-Freezing achieves sparse connections in the fully connected layer by setting the learnable parameters of the masked part to 0.
    \item \textbf{Information Propagation}: Dropout affects both forward and backward propagation, while Weight-Freezing only impacts backward propagation.
    \item \textbf{Mode of Action}: Neurons dropped by dropout will entirely lose their decision-making ability, while Weight-Freezing only influences part of the neuron's decision-making ability.
    \item \textbf{Flexibility}: Both dropout and Weight-Freezing can be applied to the fully connected layer. Nowadays, dropout can also be applied to non-fully connected layers. Moreover, because dropout and Weight-Freezing have different implementation principles, they can be used together.
\end{enumerate}

\section{Datasets}
\label{sec:datasets}

\subsection{Dataset 1: MI (BCI4-2A)}
The BCI4-2A dataset \footnote{\url{www.bbci.de/competition/iv/\#dataset2a}} was collected using a 10-20 system with 22 EEG channels, sampled at a rate of 250 Hz. The dataset includes data from nine healthy participants (ID A01-A09), across two distinct sessions. Each participant was tasked with four different motor imagery exercises: imagining the movement of the left hand, right hand, both feet, and tongue. Each session comprises 288 trials of EEG data. The data collected in the first session was used for training purposes, while the data from the second session was used for testing. Contrary to the approach of Shallow-ConvNet (\cite{schirrmeister2017deep}), who included data from 0.5s before the cue, this study only selected the time period of the motor imagery task for each trial's temporal segmentation, specifically, [2, 6] seconds after each MI cue, unless otherwise stated.

\subsection{Dataset 2: MI (BCI4-2B)}
The BCI4-2B dataset \footnote{\url{www.bbci.de/competition/iv/\#dataset2b}}  was collected using a setup with 3 EEG electrode channels, sampled at a rate of 250 Hz. This dataset includes data from nine healthy participants (ID B01-B09), across five separate acquisition sessions. Each participant was tasked with two different motor imagery exercises: imagining the movement of the left hand and the right hand. The first two sessions included 120 trials per session without feedback, while the last three sessions included 160 trials per session with a smiley face on the screen serving as feedback. As outlined in the study by \cite{miao2023lmda}, all data from the first three sessions was used for training, while the data from the last two sessions was used for testing. For our experiment, a temporal segmentation of [3, 7] seconds after each MI cue was extracted as one trial of EEG data.

\subsection{Dataset 3: ME (High-Gamma)}
The High-Gamma dataset \footnote{\url{https://gin.g-node.org/robintibor/high-gamma-dataset/src/master/data}} was collected from 14 healthy participants (6 female, 2 left-handed, average age 27$\pm$3.6 years). Each participant contributed approximately 1000 four-second trials of executed movements, divided into 13 runs. The dataset was acquired using a 128-electrode EEG setup, with 44 sensors covering the motor cortex used in the study. The EEG data was collected in a lab optimized for non-invasive detection of high-frequency movement-related EEG components. The technical setup included active electromagnetic shielding, high-resolution, low-noise amplifiers, actively shielded EEG caps, and full optical decoupling. All devices were battery powered and communicated via optic fibers. The experiment involved four classes of movements: left hand, right hand, both feet, and rest, triggered by visual cues. Each run consisted of 80 trials, with each trial lasting 4 seconds, followed by a 3 to 4 second inter-trial interval. The training set comprised approximately 880 trials from all runs except the last two, while the test set included approximately 160 trials from the final two runs. The configuration of this dataset strictly adheres to \cite{schirrmeister2017deep}, except for the data preprocessing methods.

\section{Experiments}
\label{sec:Experiments}

\subsection{Preprocessing}
Before feature extraction and classification, EEG signals underwent preprocessing, which included bandpass filtering and normalization. The preprocessing methods remained consistent with our previous work (\cite{miao2022priming, miao2023lmda}).  A bandpass filter with 200-order Blackman windows was employed to filter the raw EEG data. Subsequently, the filtered raw EEG data was segmented according to the task duration in each dataset.

The raw EEG data was bandpassed to [4, 38] Hz for MI tasks and [4, 125] Hz for the ME task. The normalization methods for both MI and ME were in line with our previous work, which included trial normalization and Euclidean alignment (\cite{he2019transfer}). The specific implementation of the preprocessing method is as follows:
\begin{align}
\mathbf{x}_{i}&=\frac{\mathbf{x}_{i}}{\max \left(\left|\mathbf{x}_{i}\right|\right)}  \\
\bar{\mathbf{R}}&=\frac{1}{N} \sum_{i=1}^{N} \mathbf{x}_{i} \mathbf{x}_{i}^{T} \\
\tilde{\mathbf{x}}_{i}&=\bar{\mathbf{R}}^{-1 / 2} \mathbf{x}_{i} 
\end{align}
Where $i$ denotes the $i$th trial of $\mathbf{x}$, $\left|\cdot\right|$ denotes taking the absolute value of the matrix, $N$ represents the number of trials.

\subsection{Experimental Environment and Parameter Settings}
All experiments were conducted using the Pytorch framework on a workstation equipped with Intel(R) Xeon(R) Gold 5117 CPUs @ 2.00 GHz and Nvidia Tesla V100 GPUs. All models were trained using AdamW (\cite{loshchilov2018fixing}) as the optimizer, with the default parameters as described, and mini batches with a size of 32.

All experimental procedures utilized the hold-out test dataset methodology. To ensure the repeatability of the experiments, a unified random number seed was used in all experiments. As data augmentation methods can influence the repeatability of results, no data augmentation methods were selected in this experiment. Similarly, cross-validation methods not only affect the repeatability of experimental results but also have a significant difference with the real use scenario of BCI, so the experimental part does not consider cross-validation. 
To enhance the reproducibility of experimental results, this paper conducted all experiments without splitting the validation set. Instead, the entire training dataset was utilized to train the ANN model for a specific number of epochs. The maximum accuracy of the test set, along with statistics such as average accuracy and median accuracy at specific epochs, were recorded throughout the training process.

It is important to mention that ConvNet (\cite{schirrmeister2017deep}) offers both shallow and deep versions for EEG decoding. However, the shallow-ConvNet has been emphasized for its advantages over the deep-ConvNet. Therefore, in our experiment, we selected the shallow-ConvNet as the experimental model (referred to as ConvNet hereafter).

\subsection{Experimental Results} \label{sec:experiment}
\subsubsection{Benchmark Algorithms}

In the experiment, we applied Weight-Freezing to three distinct ANN models and compared their decoding capabilities with several classical algorithms. FBCSP (\cite{ang2008filter}), a well-known algorithm for feature extraction in MI- and ME-EEG, was employed in this study as a representative of manual feature extraction methods. ConvNet (\cite{schirrmeister2017deep}) and EEGNet (\cite{lawhern2018eegnet}), published in 2017 and 2018 respectively, are currently highly cited ANN models in the field of EEG decoding. Sinc-ShallowNet (\cite{borra2020interpretable}), published in 2020, is a lightweight network specifically designed for EEG decoding. Despite having far fewer parameters, it achieves comparable decoding accuracy to ConvNet. DRDA (\cite{zhao2020deep}), also published in 2020, utilizes data from other participants to enhance training and achieved the highest average decoding accuracy (across participants) in BCI4-2A and BCI4-2B at that time, as described in the paper. LMDA-Net (\cite{miao2023lmda}), published in 2023, is another lightweight decoding network that surpasses DRDA in terms of average decoding accuracy (across participants) in BCI4-2A and BCI4-2B. It is worth noting that Conformer (\cite{song2023Conformer}) was mentioned as the best-performing model in BCI4-2A and BCI4-2B published before LMDA-Net in 2023. However, LMDA-Net exhibits superior average classification accuracy (across participants) in both BCI4-2A and BCI4-2B, making it an important reference for model performance in terms of decoding capabilities.

To ensure a fair comparison between ConvNet, LMDA-Net, and EEGNet under the influence of Weight-Freezing, the threshold value $t$ in Weight-Freezing remained fixed throughout the experiment. Specifically, no participant-specific tuning for the threshold value was performed. This approach maximizes fairness in the comparative analysis of different models and ensures the reproducibility of the experimental results.

\subsubsection{BCI4-2A} 

\begin{table*}[]
\caption{Classification performance of different algorithms on BCI4-2A}
\centering
\label{table:2a}
\begin{threeparttable}
\begin{tabular}{ccccccccccc}
\hline
\textbf{Methods}             & \textbf{A01}  & \textbf{A02}  & \textbf{A03}  & \textbf{A04}  & \textbf{A05}  & \textbf{A06}  & \textbf{A07}  & \textbf{A08}  & \textbf{A09}  & \textbf{mean$\pm$std} \\ \hline
FBCSP+NBPW                   & 76.0          & 56.5          & 81.2          & 61.0          & 55.0          & 45.2          & 82.7          & 81.2          & 70.7          & 67.7$\pm$13.7         \\
ConvNet                  & 76.4          & 55.2          & 89.2          & 74.6          & 56.9          & 54.1          & 92.7          & 77.1          & 76.4          & 72.5$\pm$14.2         \\
EEGNet                   & 75.3          & 51.0          & 88.5          & 57.3          & 46.5          & 50.3          & 83.7          & 80.5          & 87.1          & 68.9$\pm$17.4         \\
Sinc-ShallowNet              & -             & -             & -             & -             & -             & -             & -             & -             & -             & 72.8$\pm$12.9         \\
DRDA         & 83.1          & 55.1          & 87.4          & 75.2          & 62.2          & 57.1          & 86.1          & 86.1          & 82.0          & 74.7$\pm$13.0         \\
LMDA                     & 86.5          & 67.4          & 91.7          & 77.4          & 65.6          & 61.1          & 91.3          & 83.3          & 85.4          & 78.8$\pm$11.5         \\ \hline
EEGNet+WF(t=0.3)  & 78.8          & \textbf{70.5} & 94.8          & 63.5          & 67.0          & 55.9          & 87.8          & 82.6          & 89.9          & 76.8$\pm$13.2         \\ 
LMDA+WF(t=0.4)     & 86.8          & 69.8          & 92.7          & 81.6          & 65.6          & \textbf{68.1} & 92.7          & 84.0          & 87.8          & 81.0$\pm$10.6         \\ 
ConvNet+WF(t=0.3) & \textbf{91.3} & 66.3          & \textbf{95.5} & \textbf{85.4} & \textbf{77.8} & 66.3          & \textbf{96.9} & \textbf{91.0} & \textbf{93.2} & \textbf{84.9}$\pm$12.0         \\ \hline
\end{tabular}
    \begin{tablenotes}
        \footnotesize
        \item WF indicates the application of Weight-Freezing and t represents the threshold in Weight-Freezing.
        \item All methods using Weight-Freezing were trained for 800 epochs, and the best accuracy was recorded. The results of the remaining methods were obtained from published paper. 
        \item The specific performance of each participant in Sinc-ShallowNet is not disclosed, hence indicated as "-".
        \item Please refer to Section \ref{sec:fc-vs-wf} for the comparative experiments between Weight-Freezing and the fully connected layer.
    \end{tablenotes}
\end{threeparttable}
\end{table*}

Table \ref{table:2a} shows the classification performance of different algorithms on BCI4-2A.As observed in Table \ref{table:2a}, Weight-Freezing demonstrates a significant improvement in the classification performance of all experimental models in the BCI4-2A dataset. Particularly noteworthy are the average classification accuracies of LMDA-Net and ConvNet when utilizing Weight-Freezing, both surpassing 81\%. ConvNet with Weight-Freezing achieves an impressive average decoding accuracy of 84.9\%. To the best of our knowledge, this is the highest decoding accuracy (average across participants) currently achievable for this dataset, an increase of 10.2\% over the best-performing DRDA in 2020. Sinc-ShallowNet, EEGNet, and LMDA-Net are all lightweight networks, and from the experimental results, Weight-Freezing can not only be applied to ConvNet, which has a large number of parameters, but also to EEGNet and LMDA-Net, each with only a few thousand parameters. 
From Table \ref{table:2a}, it can also be seen that EEGNet does not have an advantage in decoding ability in this dataset, performing the worst among the chosen neural network methods. However, with Weight-Freezing, the performance of EEGNet surpasses DRDA, reaching an average decoding accuracy of 76.8\%. 
Of course, the significant improvement in classification performance of EEGNet and ConvNet is not only due to Weight-Freezing. Taking ConvNet as an example, apart from the classification network, ConvNet with Weight-Freezing demonstrates differences from the original ConvNet paper, such as variations in the number of sampling points in a trial, digital filters, preprocessing methods, and optimizers.
Among the compared models, the one most similar to our experimental setup is LMDA-Net. However, it is worth noting that the original paper of LMDA-Net conducted experiments for 300 epochs, whereas our experiments in Table \ref{table:2a} were conducted for 800 epochs. Additionally, in our experiment, the number of sampling points in each trial was reduced from the original 1125 points in LMDA-Net to 1000 points.
In Section \ref{sec:analyses}, we conducted comparative experiments to investigate the impact of Weight-Freezing and full connection on the classification accuracy of each participant. 
However, it is evident from the experimental results that EEGNet, LMDA-Net, and ConvNet exhibit significant improvements in classification performance after applying Weight-Freezing. These models outperform the known models by a considerable margin. This finding indicates that Weight-Freezing is a more suitable classification technique compared to the fully connected network. The fact that EEGNet, LMDA-Net, and ConvNet benefit from Weight-Freezing demonstrates that Weight-Freezing is a universal classification technique, effectively applicable to various types of ANNs.

\subsubsection{BCI4-2B}

\begin{table*}[]
\caption{classification performance of different algorithms on BCI4-2B}
\centering
\label{table:2b}
\begin{threeparttable}
\begin{tabular}{ccccccccccc}
\hline
\textbf{Methods}             & \textbf{B01}  & \textbf{B02}  & \textbf{B03}  & \textbf{B04}  & \textbf{B05}  & \textbf{B06}  & \textbf{B07}  & \textbf{B08}  & \textbf{B09}  & \textbf{mean$\pm$std}  \\ \hline
FBCSP+NBPW                   & 70.0          & 60.3          & 60.9          & 97.5          & 93.1          & 80.6          & 78.1          & 92.5          & 86.8          & 80.0$\pm$13.9          \\ 
ConvNet                  & 74.3          & 56.0          & 57.5          & 97.5          & 95.3          & 82.1          & 79.6          & 87.5          & 86.5          & 79.6$\pm$14.8          \\ 
EEGNet                   & 77.5          & 61.0          & 63.1          & \textbf{98.4} & \textbf{96.5} & 83.7          & 84.3          & 92.8          & 88.4          & 82.9$\pm$13.5          \\ 
DRDA         & 81.3          & 62.8          & 63.6          & 95.9          & 93.5          & 88.1          & 85.0          & 95.2          & 90.0          & 83.9$\pm$12.8          \\ 
LMDA                     & 81.6          & 63.6          & 72.8          & 98.1          & 96.3          & 90.3          & 85.3          & 95.0          & 89.7          & 85.5$\pm$12.0          \\ \hline
ConvNet+WF(t=0.7) & 80.0          & 56.1          & 61.9          & 97.8          & 92.2          & \textbf{89.4} & 84.1          & 93.1          & 88.8          & 82.6$\pm$14.4          \\ 
LMDA+WF(t=0.9)     & \textbf{82.5} & 63.2          & 71.9          & \textbf{98.4} & 94.7          & 88.1          & 85.9          & 94.4          & \textbf{92.8} & 85.8$\pm$11.5          \\ 
EEGNet+WF(t=0.8)   & 81.3          & \textbf{67.1} & \textbf{75.6} & 98.1          & 94.1          & 87.5          & \textbf{87.5} & \textbf{95.9} & 91.3          & \textbf{86.5$\pm$10.2} \\ \hline
\end{tabular}
    \begin{tablenotes}
        \footnotesize
        \item WF indicates the application of Weight-Freezing and t represents the threshold in Weight-Freezing.
        \item All methods using Weight-Freezing were trained for 800 epochs, and the best accuracy was recorded. The results of the remaining methods were obtained from published paper.
    \end{tablenotes}
\end{threeparttable}
\end{table*}

We further verified the classification performance of different algorithms on the BCI4-2B dataset. As can be seen from Table \ref{table:2b}, the classification performance of each algorithmdoes not vary as much as on BCI4-2A. In this dataset, ConvNet, which has a large number of parameters, does not have an advantage. With Weight-Freezing, ConvNet has an average classification accuracy of 82.6\% on this dataset, lower than DRDA, which has a feature extraction network similar to ConvNet. 
After applying Weight-Freezing, LMDA-Net and EEGNet achieved average classification accuracies of 85.8\% and 86.5\%, respectively. To the best of our knowledge, these results surpass the previous state-of-the-art algorithms in terms of classification performance on this dataset. 
An interesting phenomenon is that in BCI4-2B, to achieve better classification results, the threshold in Weight-Freezing is much larger than on BCI4-2A. The thresholds in Weight-Freezing for ConvNet and EEGNet are both 0.7, which means that 70\% of the weights are frozen in the fully connected layer, i.e., only a few input neurons are involved in the decision-making process for a specific motor imagery task.  This might be a compromise in situations with low signal-to-noise ratio and small amounts of data, as it's challenging to distinguish noise from real and effective features with the current limited data. In this case, letting the decision-making process for a certain category only consider a small part of the input neurons can help eliminate the interference of noise. 

Although BCI4-2B is a completely different EEG dataset compared to BCI4-2A, the classification performance of EEGNet, LMDA-Net, and ConvNet with Weight-Freezing shows improvements compared to the models being compared. This indicates that Weight-Freezing technique not only exhibits strong robustness across different ANN models but also demonstrates robustness across different EEG datasets.

\subsubsection{High-Gamma}

\begin{table*}[]
\caption{Classification performance of different algorithms on High-Gamma}
\centering
\label{table:high-gamma}
\begin{threeparttable}
\begin{tabularx}{0.7\textwidth}{>{\centering\arraybackslash}X>{\centering\arraybackslash}X}
\hline
\textbf{Methods}             & \textbf{mean$\pm$std} \\ \hline
FBCSP+rLDA                   & 86.0$\pm$9.0          \\ 
EEGNet                       & 88.5$\pm$11.0         \\ 
DeepConvNet                  & 88.4$\pm$8.8          \\ 
ConvNet                      & 93.9$\pm$9.3          \\ 
Sinc-ShallowNet              & 91.2$\pm$9.1          \\ \hline
LMDA+WF(t=0.3)     & 93.6$\pm$5.8          \\ 
EEGNet+WF(t=0.3)  & 94.0$\pm$4.6          \\ 
ConvNet+WF(t=0.5) & \textbf{96.9}$\pm$2.4          \\ \hline
\end{tabularx}
    \begin{tablenotes}
        \footnotesize
        \item WF indicates the application of Weight-Freezing and t represents the threshold in Weight-Freezing.
        \item All methods using Weight-Freezing were trained for 800 epochs, and the best accuracy was recorded. The results of the remaining methods were obtained from published paper.
    \end{tablenotes}
\end{threeparttable}
\end{table*}

Weight-Freezing has demonstrated outstanding performance on the aforementioned motor imagery datasets. This success can be attributed to the non-stationary nature and low signal-to-noise ratio of motor imagery signals. We further assessed the classification performance of Weight-Freezing in the context of motor execution EEG (ME-EEG). The High-Gamma dataset comprises four-class motor execution tasks, with each participant's training data volume significantly larger than that of BCI4-2A and BCI4-2B. As no existing work has provided a detailed classification performance for each participant on the High-Gamma dataset, we only present the average accuracy and standard deviation across all participants. For a detailed analysis of each participant's classification performance, please refer to the Appendix.

From Table \ref{table:high-gamma}, we can see that ME decoding is much easier than MI decoding. FBCSP+rLDA achieves an average accuracy of 86\% for a four-classification task. Weight-Freezing also works on High-Gamma, with LMDA-Net, EEGNet, and ConvNet all surpassing 93\% classification accuracy under the boost of Weight-Freezing. The performance of ConvNet reached 96.9\%, with Weight-Freezing. To the best of our knowledge, this is also the highest classification accuracy that can currently be achieved on this dataset. This suggests that Weight-Freezing remains effective on ME-EEG  with a larger number of training samples. This yet again confirms that Weight-Freezing is a suitable classification network for EEG decoding.

\section{Analyses}
\label{sec:analyses}

In this section, we conducted an in-depth analysis of Weight-Freezing using the challenging BCI4-2A dataset as a case study. We employed a controlled variable approach to investigate the performance differences before and after applying Weight-Freezing in three models: ConvNet, EEGNet, and LMDA-Net. In section \ref{sec:fc-vs-wf}, we provided a comprehensive comparison of the performance curves between different algorithms before and after applying Weight-Freezing, shedding light on the test accuracy and training epochs under different algorithms for each participant. In section \ref{sec:Robust-length}, we examined the robustness of Weight-Freezing with respect to the number of sampling points in the input EEG data. Moreover, in section \ref{sec:robustness-volume}, we expanded the number of EEG samples and employed transfer learning to validate the robustness of Weight-Freezing with large sample volumes. Finally, in section \ref{threshold}, we analyzed the impact of the threshold in Weight-Freezing on accuracy.

\subsection{Training Process Analysis Before and After Weight-Freezing} \label{sec:fc-vs-wf}
\begin{figure*}
\centerline{\includegraphics[width=\textwidth]{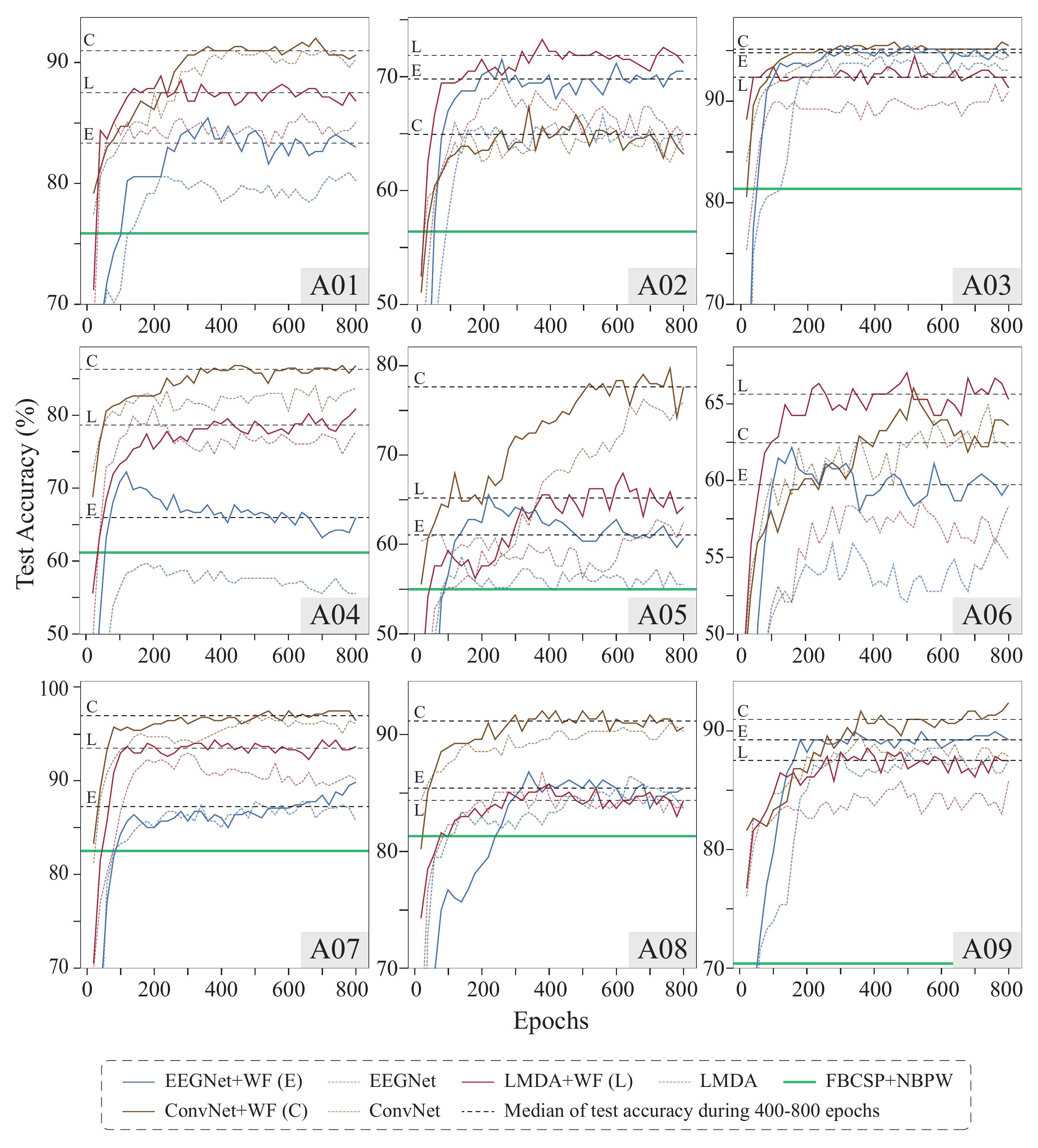}}
\caption{Comparison of the training processes before and after Weight-Freezing for EEGNet, ConvNet, and LMDA on  BCI4-2A. To provide a clear representation of the test results for different models, a smoothing window of width 20 was applied to the test accuracy of the ANN models. The figure also shows the median test accuracy of the aforementioned models from the 400th to the 800th epoch after applying Weight-Freezing. FBCSP+NBPW achieved a classification accuracy of less than 50\% in A06 and is therefore not displayed.}
\label{fig3}
\end{figure*}

Although section \ref{sec:experiment} presented the test accuracy of different algorithms after applying Weight-Freezing, it only shows the superiority of our method in decoding, and cannot specifically reflect the changes in the training process before and after Weight-Freezing. Therefore, in this subsection, we fixed all experimental conditions to compare in detail the differences between different algorithms before and after applying Weight-Freezing. 
As can be seen from Figure \ref{fig3}, for almost every participant and every algorithm, the decoding performance improved significantly after applying Weight-Freezing. This improvement in decoding accuracy is not due to large fluctuations, but because ANN models indeed received effective training. This indicates that, under the same number of training epochs, ANN models exhibit better performance after applying Weight-Freezing.  Moreover, the observed results depicted in Figure \ref{fig3} are consistent across almost all participants and all tested ANN models.

The large fluctuations of ANN models during the training process is a significant reason many scholars still prefer FBCSP as a feature extraction method. Due to the small amount of training data and the non-stationary nature of EEG, if the training data is re-divided into training and validation sets, not only will the performance of the model be reduced due to the reduction in training set data, but the distribution of data in the validation set is likely to differ greatly from the test data distribution. This means that using an early-stopping strategy in EEG decoding is extremely difficult, because the loss and accuracy of the ANN model in the validation set do not intuitively reflect the situation in the test set. If the ANN model can achieve good and stable performance by training for a specific number of epochs, this will be  highly valuable for the practical application of BCI systems.
As can be seen from Figure \ref{fig3}, after about 200 training epochs, the test accuracy of different models in all participants far exceeds that of FBCSP+NBPW. ConvNet and LMDA-Net even need less than 100 epochs of training to far exceed the classification performance of FBCSP+NBPW. From the median test accuracy of different models during the 400-800 epochs, it can be seen that ANN models under the boost of Weight-Freezing have already opened a huge gap in the classification performance of all participants compared to FBCSP+NBPW, thus proving that the Weight-Freezing technology is a key technology that promises to enhance the decoding ability of ANN.

\subsection{Robustness of Sampling Time Length in an EEG Trial} \label{sec:Robust-length}
In our previous research (\cite{miao2022priming, miao2023lmda}), we have always assumed that the time-domain data truncation method proposed by \cite{schirrmeister2017deep} is optimal, i.e., truncating time-domain data starting from 0.5s before motor imagery or execution. In this subsection, we validated the robustness of ANN models to different lengths of time-domain data in each trial before and after Weight-Freezing. In this experiment, we focus on the influence of the duration of the task period on the test results, without considering the variation in data length caused by down-sampling the sampling frequency. 

\begin{figure*}
\centerline{\includegraphics[width=\textwidth]{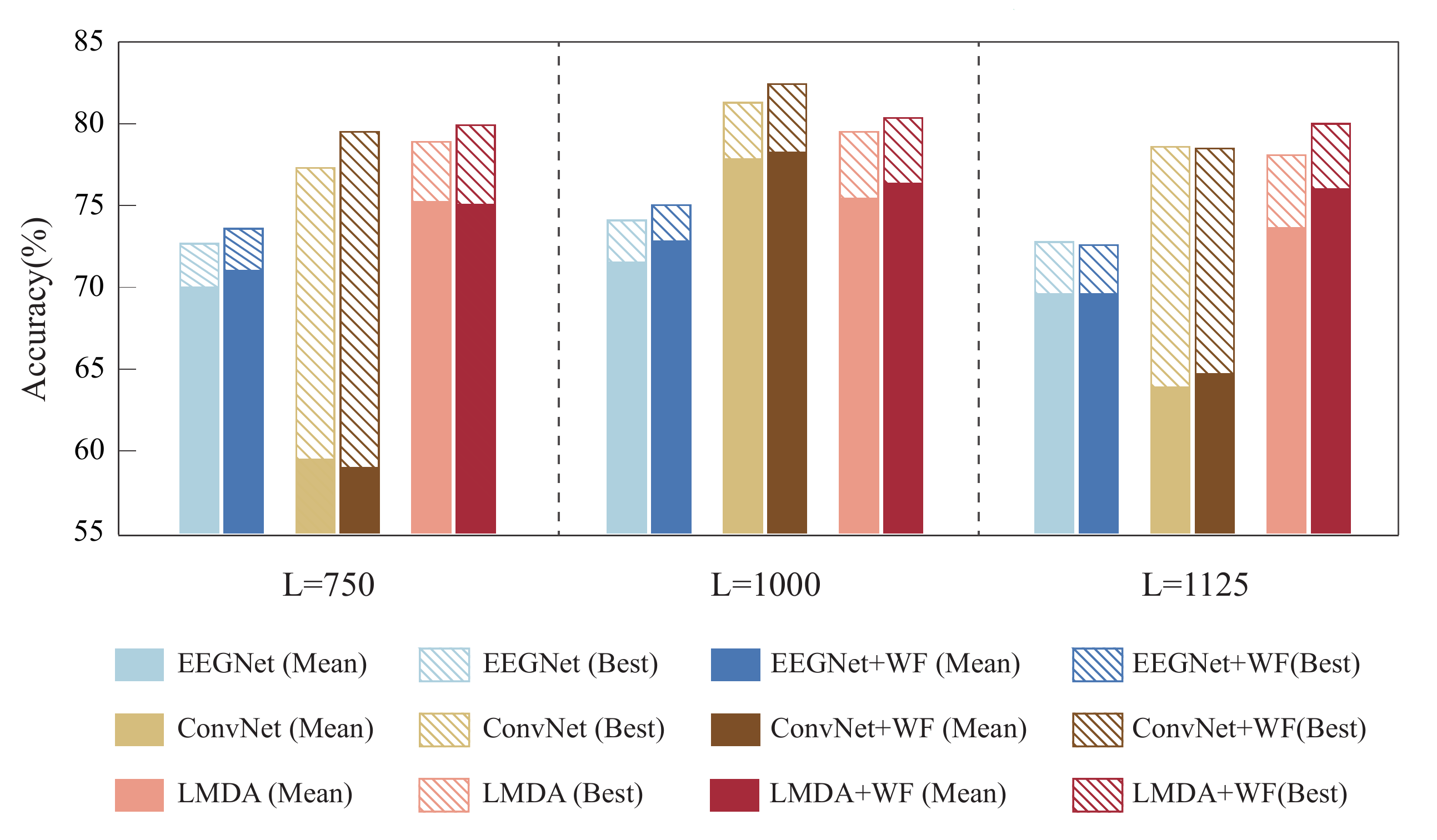}}
\caption{Robustness analysis of ANN models to Sampling Time Length in an EEG Trial before and after applying Weight-Freezing on BCI4-2A. All ANN models were trained for 300 epochs, and the average accuracy of all participants was recorded. The highest accuracy of ANN models is shown by a shaded column, while the mean accuracy throughout the last 10 epochs is shown by an opaque solid column.(Best viewed in color)}
\label{fig4}
\end{figure*}

In Figure \ref{fig4}, L=1000 indicates that the complete motor imagery task period of 4s data is selected for each trial, L=750 indicates that only the first three seconds of data during the motor imagery task period is selected for each trial, and L=1125 indicates that not only the complete motor imagery task period data is included for each trial, but also the data from 0.5s before the task period.
As can be seen from Figure \ref{fig4}, EEGNet, ConvNet, and LMDA-Net have almost all reached the highest average accuracy after applying Weight-Freezing, indicating that Weight-Freezing exhibits good robustness in this scenario and can be used as a universal classification network. However, in the last 10 epochs of average accuracy, applying Weight-Freezing does not bring a significant improvement to the test fluctuation of the ANN model, especially for ConvNet. The fluctuations observed in ConvNet during the training process are not related to Weight-Freezing but are primarily influenced by the network architecture itself. From Figure \ref{fig4}, it can be observed that the classification performance of LMDA-Net and EEGNet exhibits less fluctuation than ConvNet across different scenarios. Weight-Freezing, as a classification network, can enhance the decoding capability of ANN models, but its ability to improve the prediction fluctuation of the models is limited. The prediction fluctuation of ANN models is primarily determined by the network architecture itself.

If the complete motor imagery task period is used as the benchmark, that is, L=1000 is the standard trial length, L=750 and L=1125 reflect two different scenarios: missing time-domain data and adding noise to the time-domain data. The classification performance of different algorithms under these three different scenarios is worth deep discussion. In these three different scenarios, ConvNet can achieve high classification accuracy, but its prediction fluctuation for different scenarios is also the largest. This indicates small disturbances for the parameters in ConvNet can bring large differences to the decision results, which is an undesired feature in EEG decoding. A possible reason is that ConvNet has too many learnable parameters, and only a few hundred training samples cannot effectively train them. Although the decoding ability of ConvNet has been slightly improved after applying Weight-Freezing, it is still unable to cope with the fluctuations in the feature extraction network during the training process. Lightweight networks like EEGNet and LMDA-Net, have relatively stable classification performance under these three scenarios, and the introduction of Weight-Freezing also improves their decoding ability to some extent. The implementation of the lightweight network itself incorporates a strong regularization approach, and the utilization of Weight-Freezing can be considered as an additional regularization process. This highlights the significance of regularization techniques in EEG decoding. Comparing the performance of different models in the scenarios of L=750 and L=1125, it is found that at L=750, even though 25\% of the effective data in the task period is discarded, the classification performance of the model is not lower than its performance at L=1125. Compared to the scenario of L=1125, the scenario of L=750 itself is also a kind of regularization. The classification performance of different models at L=750 also raises a question for the designers of motor imagery experiments: do we really need such a long task period in motor imagery experiment?

\subsection{Robustness to the Volume of Training Data} \label{sec:robustness-volume}

\begin{table*}[]
\caption{Classification performance of different algorithms on BCI4-2A dataset under transfer learning}
\centering
\label{table:transfer}
\begin{threeparttable}
\begin{tabular}{ccccccccccc}
\hline
\textbf{Methods}              & \textbf{A01}  & \textbf{A02}  & \textbf{A03}  & \textbf{A04}  & \textbf{A05}  & \textbf{A06}  & \textbf{A07}  & \textbf{A08}  & \textbf{A09}  & \textbf{mean+std}  \\ \hline
DRDA                          & 69.5          & 26.7          & 82.5          & 41.3          & 27.6          & 32.4          & 49.6          & 76.5          & 70.1          & 52.9+22.1          \\ 
EEGNet                        & 77.1          & \textbf{47.9} & 78.1          & 45.8          & 44.8          & \textbf{52.8} & 66.3          & 63.9          & 63.9          & 60.1+12.8          \\ 
ConvNet                       & \textbf{82.6} & 42.0          & 85.8          & 53.1          & 48.3          & 49.7          & 51.0          & \textbf{77.8} & \textbf{75.7} & 62.9+17.1          \\ 
LMDA                          & 78.5          & 44.4          & 80.6          & 49.3          & 46.2          & 47.9          & 56.3          & 76.7          & 73.3          & 61.5+15.4          \\ \hline
EEGNet+WF(t=0.5)  & 78.8          & 45.1          & 83.3          & \textbf{53.8} & 46.9          & 49.7          & \textbf{68.1} & 69.8          & 66.3          & 62.4+14.1          \\ 
LMDA+WF(t=0.2)    & 77.4          & 42.0          & 80.9          & 52.4          & 50.7          & 52.1          & 58.7          & \textbf{77.8} & 75.3          & 63.0+14.8          \\ 
ConvNet+WF(t=0.2) & 81.9          & 41.3          & \textbf{84.4} & 54.9          & \textbf{51.7} & 51.0          & 53.1          & \textbf{77.8} & \textbf{75.7} & \textbf{63.5}+16.2 \\ \hline
\end{tabular}
    \begin{tablenotes}
        \footnotesize
        \item WF indicates the application of Weight-Freezing and t represents the threshold in Weight-Freezing.
        \item Except for DRDA, whose results were disclosed in the published paper, all other methods using Weight-Freezing were trained for 300 epochs, and the best accuracy was recorded.
    \end{tablenotes}
\end{threeparttable}
\end{table*}

The lack of training data is a significant challenge in EEG decoding. On BCI4-2A, each participant only has 288 training samples, while the number of parameters in EEGNet, LMDA-Net, and ConvNet are 3012, 4284, and 47364, respectively. ANNs can easily overfit under such a small number of training samples. Therefore, implementing regularization on ANNs is crucial, which is a significant reason why Weight-Freezing is effective. However, one question arises: what if we have ample training samples? Will Weight-Freezing still be effective? Unfortunately, currently, there is no publicly available EEG dataset with tens of thousands of samples. For this, we have two ways to augment the training samples. The first is data augmentation, and the second is transfer learning. Considering the possibility of discrepancies and uncertainties between artificially generated EEG data and real signals, we adopt transfer learning to examine the impact of  training data volume on Weight-Freezing.

The test based on transfer learning can be considered as unsupervised learning. We only selected the test data from the target participant and used all other participants' data for training. That is, if each participant has 288 training and testing data, the number of experimental test data in this section remains 288, but the training data is $288\times2\times8$, totaling 4608. This data volume barely matches the parameter volume of EEGNet and LMDA-Net. Since the \cite{zhao2020deep} also conducted experiments in the same scenario, Table \ref{table:transfer} also shows the test accuracy of DRDA. As can be seen from Table \ref{table:transfer}, the average classification accuracy of EEGNet, ConvNet, and LMDA-Net far exceeds DRDA, indicating that it is meaningless to use the GAN (\cite{goodfellow2020generative}) method to reduce the data distribution difference between the target participant and other participants in unsupervised learning. Although the training data volume in unsupervised learning is comparable to the parameters of EEGNet and LMDA-Net,all the models do not have an obvious advantage in classification performance. This may be attributed to the fact that real EEG signals are non-stationary and characterized by a low signal-to-noise ratio. Consequently, the noise in the EEG signals inevitably impacts the decoding accuracy.It can also be seen from Table \ref{table:transfer} that Weight-Freezing can somewhat enhance the classification performance of ANNs, indicating that it is also applicable in transfer learning scenarios with a larger volume of training data.

\subsection{The Impact of the Threshold in Weight-Freezing} \label{threshold}
\begin{figure*}
\centerline{\includegraphics[width=0.8\textwidth]{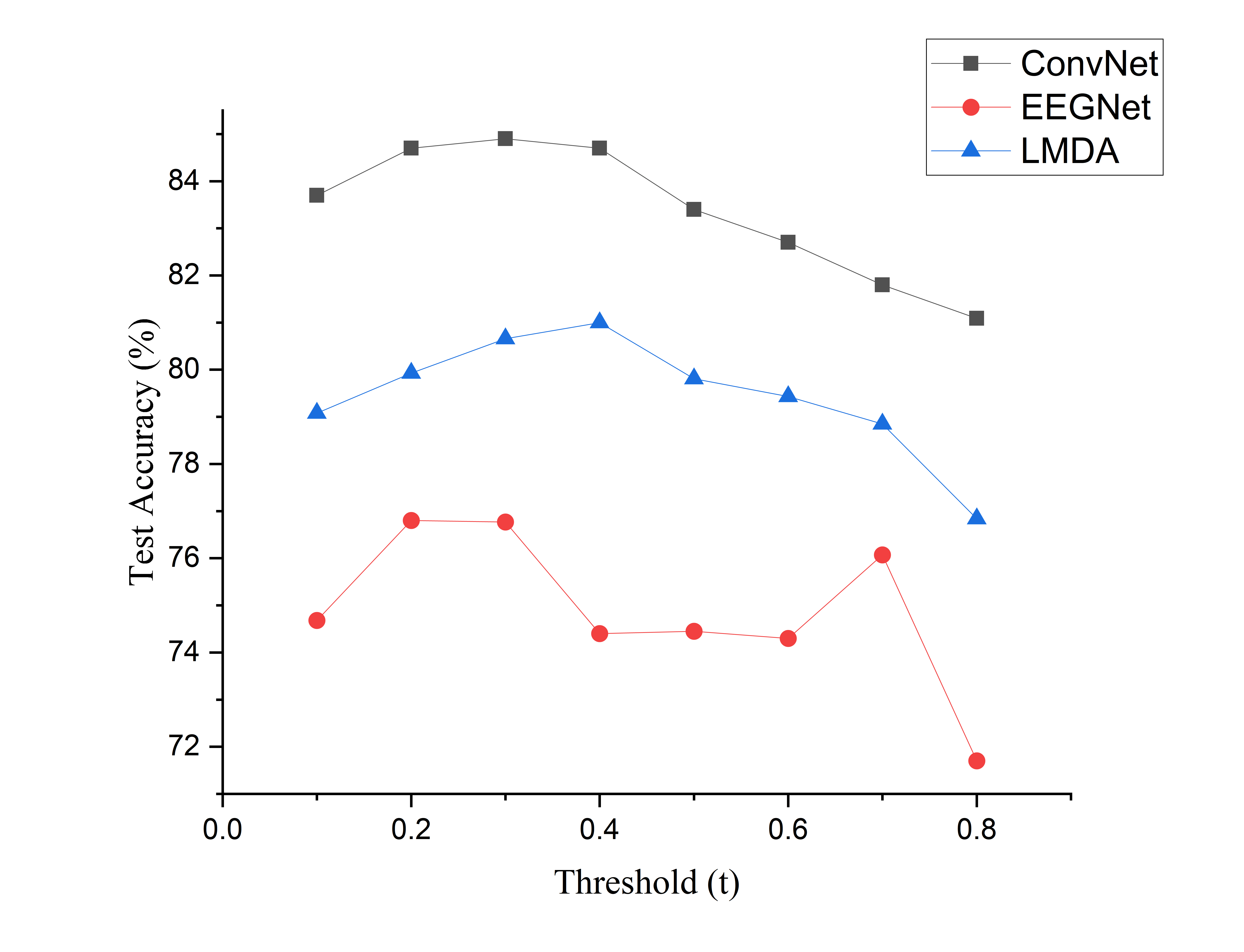}}
\caption{The impact of the threshold in Weight-Freezing of different ANN models on BCI4-2A. All models were trained for 800 epochs, and the average accuracy for all participants was recorded.}
\label{fig5}
\end{figure*}

The threshold in Weight-Freezing represents the proportion of frozen parameters in the fully connected layer. The larger the threshold, the more parameters in the fully connected layer are frozen. Hence, we further validated the impact of the threshold value t on the classification results of different ANN models. We chose different threshold values t from the set \{0.1, 0.2, 0.3, 0.4, 0.5, 0.6, 0.7, 0.8\}, respectively trained ConvNet, EEGNet, and LMDA-Net (run for 800 epochs), and plotted the average accuracy of all participants under different threshold values for each model, as shown in Figure \ref{fig5}. 

As can be seen from Figure \ref{fig5}, LMDA-Net and ConvNet have similar trends in sensitivity to the threshold, and EEGNet also has a similar trend except for t=0.7. For ConvNet, when t=0.3, it has the best classification performance, with an average accuracy of 84.9\% on BCI4-2A. For LMDA-Net, when t=0.4, it has the best classification performance, with an average accuracy of 81.0\%, while EEGNet has the same classification performance at t=0.2 and 0.3, both at 76.8\%. From Figure \ref{fig5}, it can also be seen that a larger threshold will affect the model's classification performance. The classification accuracy of ConvNet and LMDA-Net has significantly declined when t$>$0.5, and the classification accuracy of EEGNet has also significantly decreased when t$>$0.4 (excluding outlier t=0.7). The impact of the threshold t on the classification result also complies with the prior knowledge of neuroscience, that is, not all input neurons need to be considered in the decision process of a specific category. In the decision process of a particular category, neglecting the impact of some neurons is meaningful for EEG feature classification.  Choosing an appropriate threshold according to the ANN architecture and decoding task can fully exploit the advantages of different decoding networks, thereby improving ANN's classification performance for EEG.

\section{Conclusion and outlook}
\label{sec:Conclusion-outlook}

\subsection{Conclusion}
\label{sec:conclusion}

In this paper, we propose a novel approach to enhance the classification performance of ANNs for EEG decoding. Our approach, called Weight-Freezing, is based on the principles of ANN regularization and neuroscience prior knowledge. Weight-Freezing involves freezing specific weights in the fully connected layer by utilizing a mask matrix and a threshold, thus reducing the influence of certain neurons on the decision-making process for a particular EEG task. By setting the masked weights to zero, Weight-Freezing achieves sparse connections in the fully connected layer, effectively serving as a regularization method.
We conducted experiments using three different ANN architectures and three highly cited EEG datasets. The results demonstrate that Weight-Freezing significantly improves the classification performance of the three ANN models and surpasses the previous highest average accuracy achieved on the three datasets. In the analysis section, we conducted control experiments to compare the differences before and after Weight-Freezing for different models and investigated the impact of the threshold in Weight-Freezing on the classification results.
Our findings highlight the superiority of Weight-Freezing as a network architecture for EEG feature classification compared to the traditional fully connected network. This approach has the potential to make significant contributions in future research endeavors.

\subsection{Outlook}
\label{sec:outlook}

This paper has introduced Weight-Freezing as a regularization method for the fully connected layer in the context of EEG decoding. Although the current research has primarily focused on its effectiveness in EEG decoding, we believe that Weight-Freezing has broader applications beyond this domain. Weight-Freezing, as a universal regularization technique, enables sparse connections in the fully connected layer, which sets it apart from traditional methods like Dropout.  We postulate that its application could extend beyond EEG decoding and make significant contributions in other fields.
Moreover, our research has validated the performance of Weight-Freezing within the specific confines of ANN networks, primarily those based on convolutional networks. Given the burgeoning success of transformer networks in domains such as machine vision and natural language processing, we posit that the exploration of Weight-Freezing in transformer networks' fully connected layers represents a promising avenue for future research.
Our future work will aim to extend the current study by investigating the performance of Weight-Freezing in EEG decoding based on transformer models. We believe that this direction holds considerable potential to further enhance our understanding of Weight-Freezing's capabilities and broaden its application.

\section*{Appendix}
The classification performance of ConvNet, EEGNet, and LMDA-Net, after applying Weight-Freezing, on the High-Gamma dataset for each participant is presented in Table \ref{table_highGamma_details}. 

\begin{table*}[]
\caption{Classification performance of different algorithms on High-Gamma}
\centering
\label{table_highGamma_details}
\begin{threeparttable}
\begin{adjustbox}{width=1\textwidth}
\begin{tabular}{ccccccccccccccc}
\hline
\textbf{Methods}               & \textbf{S01} & \textbf{S02} & \textbf{S03} & \textbf{S04} & \textbf{S05} & \textbf{S06} & \textbf{S07} & \textbf{S08} & \textbf{S09} & \textbf{S10} & \textbf{S11} & \textbf{S12} & \textbf{S13} & \textbf{S14} \\ \hline
LMDA+WF (t=0.3)    & 96.9         & 90.0         & 99.4         & 100.0        & 96.3         & 90.0         & 92.5         & 91.9         & 98.1         & 92.5         & 76.9         & 95.6         & 93.1         & 96.9         \\ 
EEGNet+WF (t=0.3)  & 90.6         & 93.8         & 100.0        & 99.4         & 95.0         & 93.8         & 91.8         & 96.3         & 99.4         & 89.4         & 91.9         & 96.9         & 95.6         & 83.1         \\ 
ConvNet+WF (t=0.5) & 95.0         & 95.6         & 100.0        & 100.0        & 98.8         & 96.9         & 94.3         & 95.0         & 98.8         & 95.6         & 98.1         & 98.1         & 98.7         & 91.9         \\ \hline
\end{tabular}
\end{adjustbox}
    \begin{tablenotes}
        \footnotesize
        \item WF indicates the application of Weight-Freezing and t represents the threshold in Weight-Freezing.
        \item All ANN models were trained for 800 epochs, and the best accuracy was recorded.
    \end{tablenotes}
\end{threeparttable}
\end{table*}

\section*{Statistics and reproducibility}  
We used the hold-out test set method in all experiments and fixed all initialization seeds to ensure the reproducibility of the results obtained from the neural network model.
The hyperparameters used in this study were not optimized for each participant individually, but were instead set based on the average accuracy of all participants. Specifically, the same initialization parameter was applied to all participants for each classification scenario.
Given the large individual differences observed in our study, we computed the average accuracy and variance for all participants. 

\section*{Data and code availability statement}
The raw EEG data that support the findings of this study has been annotated  in the main text. Please refer to the footnote provided.

The source code for EEGNet , ConvNet and LMDA-Net is publically available at the following webpage: \url{https://github.com/vlawhern/arl-eegmodels}, \url{https://github.com/TNTLFreiburg/braindecode} and \url{https://github.com/MiaoZhengQing/LMDA-Code}, respectively. We are pleased to provide publiclly source code for Weight-Freezing at \url{https://github.com/MiaoZhengQing/WeightFreezing}.

\section*{Credit authorship contribution statement}
\textbf{Zhengqing Miao}: Conceptualization, Methodology, Formal analysis, Investigation, Visualization, Writing -original draft  \textbf{Meirong Zhao} Supervision, Funding acquisition. 

\section*{Competing Interests statement}
The authors declare no competing interests. 

\section*{Acknowledgments}
We extend our gratitude to Yiwei Yang for polishing the figures. We would also like to express our heartfelt appreciation to the anonymous reviewers for their constructive feedback and valuable suggestions, which significantly contributed to the improvement of the quality of our paper.

\bibliographystyle{cas-model2-names}
\bibliography{WeightFreezing_references}

\end{document}